\documentclass[runningheads]{llncs}
\usepackage{graphicx}
\usepackage{hyperref}
\usepackage{amsmath}
\usepackage{amssymb}

\usepackage{array}
\usepackage{booktabs}
\usepackage{multirow}
\usepackage{makecell}
\usepackage{ulem}
\usepackage[misc]{ifsym}

\newcommand{\equalcontribution}{\textsuperscript{*}}

\begin{document}

\title{EndoGaussian: Real-time Gaussian Splatting for Dynamic Endoscopic Scene Reconstruction}
\titlerunning{EndoGaussian for Real-time Endoscopic Scene Reconstruction}
\authorrunning{Liu et al.}
\author{Yifan Liu\inst{1} \equalcontribution \and
Chenxin Li\inst{1} \equalcontribution\and
Chen Yang\inst{2}\and
Yixuan Yuan\inst{1} $^{(\textrm{\Letter})}$
}
\institute{The Chinese University of Hong Kong\\
\email{yfliu@link.cuhk.edu.hk, yxyuan@ee.cuhk.edu.hk}\and
City University of Hong Kong
}
\maketitle % typeset the header of the contribution

\begin{abstract}
Reconstructing deformable tissues from endoscopic videos is essential in many downstream surgical applications. However, existing methods suffer from slow rendering speed, greatly limiting their practical use. In this paper, we introduce EndoGaussian, a real-time endoscopic scene reconstruction framework built on 3D Gaussian Splatting (3DGS). 
By integrating the efficient Gaussian representation and highly-optimized rendering engine, our framework significantly boosts the rendering speed to a real-time level.
To adapt 3DGS for endoscopic scenes, we propose two strategies, Holistic Gaussian Initialization (HGI) and Spatio-temporal Gaussian Tracking (SGT), to handle the non-trivial Gaussian initialization and tissue deformation problems, respectively. In HGI, we leverage recent depth estimation models to predict depth maps of input binocular/monocular image sequences, based on which pixels are re-projected and combined for holistic initialization. In SPT, we propose to model surface dynamics using a deformation field, which is composed of an efficient encoding voxel and a lightweight deformation decoder, allowing for Gaussian tracking with minor training and rendering burden.
Experiments on public datasets demonstrate our efficacy against prior SOTAs in many aspects, including better rendering speed (195 FPS real-time, 100$\times$ gain), better rendering quality (37.848 PSNR), and less training overhead (within 2 min/scene), showing significant promise for intraoperative surgery applications. 
Code is available at: \url{https://yifliu3.github.io/EndoGaussian/}.

\end{abstract}
\section{Introduction}
Reconstructing surgical scenes from endoscopic videos is crucial to robotic-assisted minimally invasive surgery (RAMIS) \cite{zha2023endosurf}. 
By recovering a 3D model of the observed tissues, such techniques facilitate simulating the surgical environment for preoperative planning and AR/VR medics training \cite{liu2020reconstructing,tang2018augmented}. Moreover, the reconstruction that supports real-time rendering can further expand its applicability to intraoperative use \cite{chen2018slam,penza2017envisors}, empowering surgeons with a complete view of the scene and facilitating their navigation and control of surgical instruments, and potentially paving the way for robotic surgery automation.

Pilot study for surgical scene reconstruction leverages depth estimation~\cite{brandao2021hapnet,luo2022unsupervised}, point cloud fusion in a SLAM-style~\cite{song2017dynamic,zhou2019real,zhou2021emdq}, and
integrating wrap fields~\cite{li2020super,long2021dssr,gao2019surfelwarp}.
With the emergence of Neural Radiance Fields (NeRFs) \cite{mildenhall2021nerf}, more recent efforts are devoted to representing the surgical scene as the radiance field
\cite{batlle2023lightneus,wang2022neural,zha2023endosurf,yang2023neural}. As a pioneer work, EndoNeRF \cite{wang2022neural} models the dynamic surgical scene as a canonical field and a time-dependent displacement field, successfully reconstructing deformable tissues. To further improve the surface reconstruction quality, EndoSurf \cite{zha2023endosurf} utilizes the signed distance field (SDF) \cite{wang2021neus,yariv2020multiview} to explicitly constrain the surface geometry. Meanwhile, Lerplane \cite{yang2023neural} treats dynamic scenes as 4D volumes and factorizes them into several explicit 2D planes, greatly accelerating the training speed. Though achieving decent results, these methods typically require querying the radiance fields repeatedly at a huge number of points and rays for rendering each image, which significantly limits their rendering speed \cite{chen2024survey} and poses great obstacles for practical applications like intraoperative use.

\begin{figure}[t]
 \centering
 \includegraphics[width=1.0\linewidth]{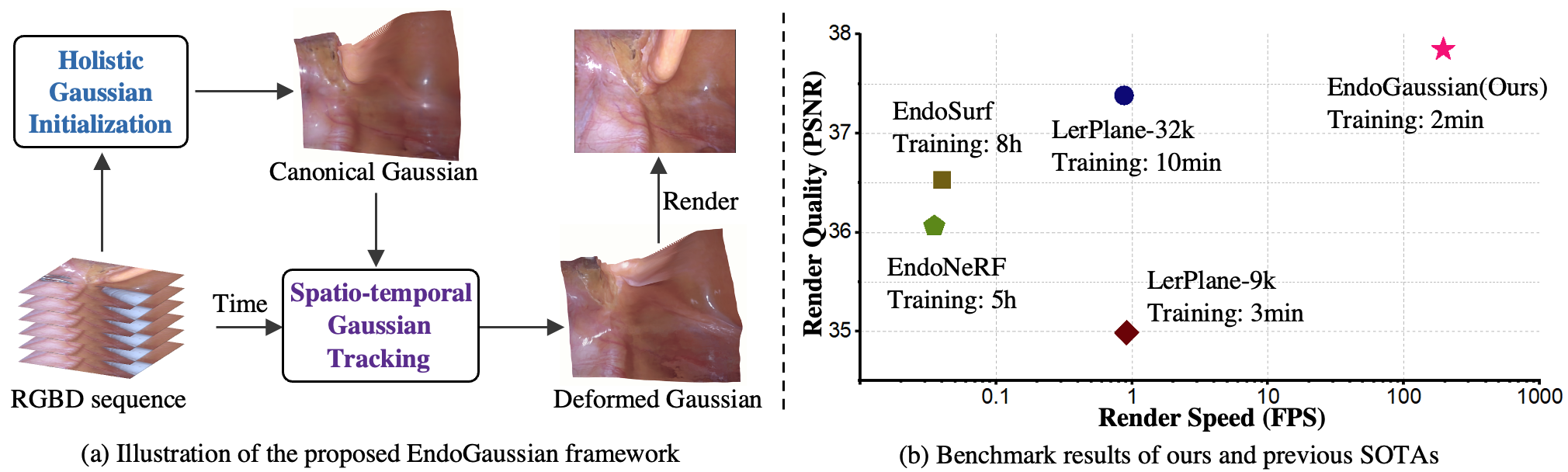}
 \caption{Illustration of (a) the pipeline of our EndoGaussian framework and (b) benchmarked results of ours against previous SOTAs on ENDONERF dataset \cite{wang2022neural}.}
 \label{fig:intro}
\end{figure}

Addressing the issue of NeRF, 3D Gaussian Splatting (3DGS) \cite{kerbl20233d} emerges as a promising alternative. By representing the scene as anisotropic 3D Gaussians and rendering images with the efficient tile-based rasterizer, it allows for real-time rendering and also superior reconstruction quality. Nevertheless, adopting 3DGS for surgical scenes is nontrivial due to two significant challenges. Firstly, 3DGS relies on Structure-from-Motion (SfM) algorithms like COLMAP \cite{schonberger2016structure} to initialize Gaussian positions. However, it is a time-consuming pipeline with multiple stages and can only produce sparse initialized points, which would hinder the optimization of 3D Gaussians due to the insufficient distribution prior \cite{zhu2023fsgs}. Secondly, the design of the original 3DGS can not handle the modeling of deformable tissues, while these tissues are prevalent during surgical procedures. 

To tackle these challenges, we propose a novel reconstruction framework named EndoGaussian, which represents the first effort to adapt 3DGS for endoscopic scene reconstruction. As shown in Fig. \ref{fig:intro} (a), we propose two novel regimes, i.e.,  \textit{Holistic Gaussian Initialization} (HGI) and \textit{Spatio-temporal Gaussian Tracking} (SGT), to initialize dense Gaussians and model surface dynamics, respectively.  The main contributions are
summarized as: (1) To achieve a fast and dense initialization in HGI, we leverage recent depth estimation models to predict absolute/relative depth values for the input binocular/monocular image sequence. Based on the predicted depth maps, pixels of input images are re-projected and combined for a holistic Gaussian initialization. (2) To model scene dynamics in SGT, we design the deformation field as a combination of efficient encoding voxel and a lightweight deformation decoder, allowing for Gaussian tracking with a minor training and rendering burden. (3) Extensive benchmark results in Fig. \ref{fig:intro} (b) on public datasets demonstrate our efficacy against prior SOTAs in many aspects, including real-time rendering efficacy (195 FPS, 100$\times$ gain), better rendering quality (37.8 PSNR), and less training overhead (within 2 min/scene), paving the way for real-time intraoperative applications.

\section{Method}
\begin{figure}[t]
 \centering
  \includegraphics[width=1.0\linewidth]{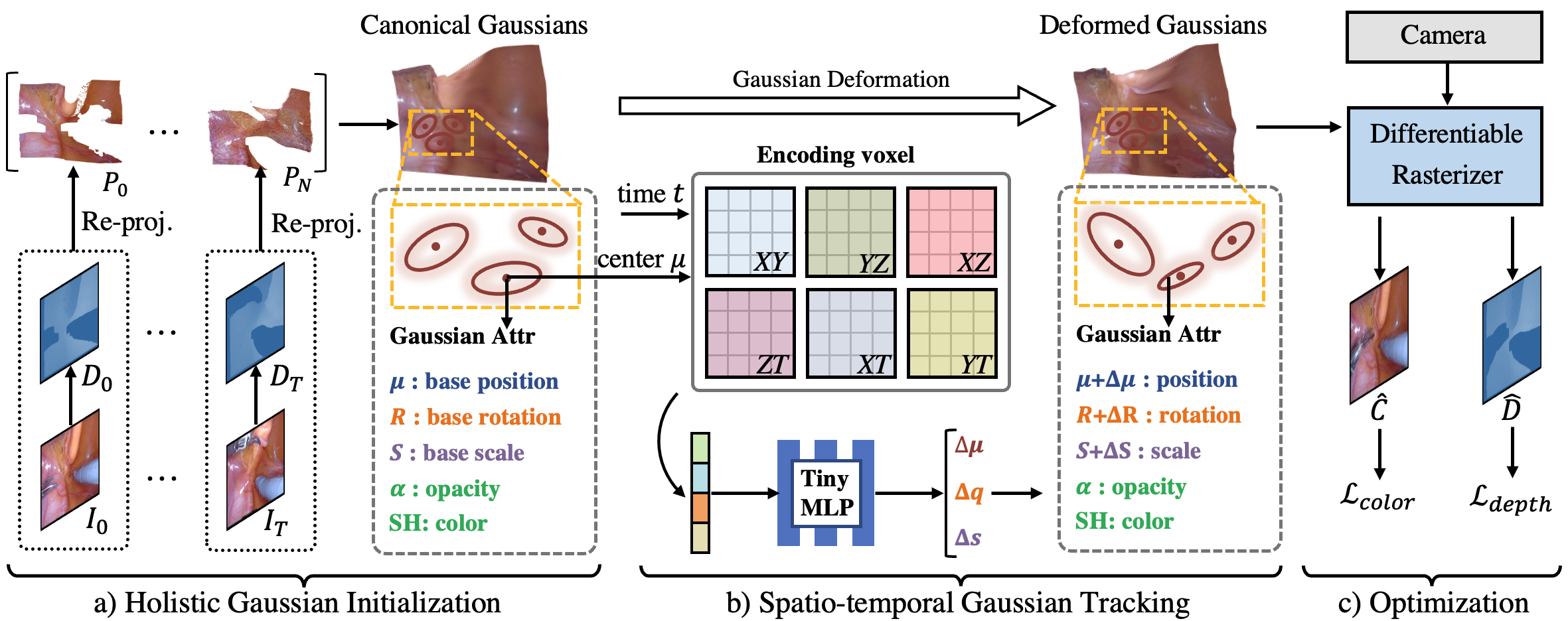}
 \caption{Illustration of the proposed EndoGaussian framework, including a) Holistic Gaussian Initialization, b) Spatio-temporal Gaussian Tracking, and c) Optimization.}
 \label{fig:framework}
\end{figure}

Our framework is designed for reconstructing surgical scenes with deformable tissues, by leveraging the recent 3D Gaussian Splatting technique (Sec. \ref{preliminaries}). As shown in Fig. \ref{fig:framework}, it begins with Holistic Gaussian Initialization to represent the scene as a set of anisotropic Gaussians with optimizable attributes (Sec. \ref{initialization}). Then, Spatio-temporal Gaussian Tracking is used to track the deformation of each Gaussian, obtaining the deformed Gaussians for a query time (Sec. \ref{tracking}). After that, differential splatting is used to render the predicted image and depth from deformed Gaussians, from which rendering and spatio-temporal constraints are computed to optimize the whole framework (Sec. \ref{optimization}).

\subsection{Preliminaries of 3DGS}
\label{preliminaries}
Gaussian splatting \cite{kerbl20233d} uses a 3D Gaussian representation to model static scenes as they can be easily projected to 2D splats, allowing fast $\alpha$-blending for image rendering. The 3D Gaussians are defined by the covariance matrix $\boldsymbol{\Sigma}$ in world space centered at the mean $\boldsymbol{\mu}$, described as $G(\mathbf{x})=\text{exp}(-1/2(\mathbf{x}-\boldsymbol{\mu})^{T}\boldsymbol{\Sigma}^{-1}(\mathbf{x}-\boldsymbol{\mu}))$, where $\boldsymbol{\Sigma}$ is decomposed into rotation matrix $\mathbf{R}$ and scaling matrix $\mathbf{S}$. To represent a scene, 3DGS creates a dense set of 3D Gaussians and optimizes their render-related attributes including positions $\boldsymbol{\mu}$, rotation $\mathbf{R}$, scaling $\mathbf{S}$, opacity $o$, and their spherical harmonic (SH) coefficients. From these attributes, the color $\mathbf{\hat{C}(x)}$ and depth $\mathbf{D(x)}$ of a certain pixel $\mathbf{x}$ can be rendered by the function:
\begin{equation}
\label{Eq:render}
    \mathbf{\hat{C}(x)}=\sum_{i=1}^{n}\mathbf{c_i}\alpha_i\prod_{j=1}^{i-1}(1-\alpha_j), \text{  } \mathbf{\hat{D}(x)}=\sum_{i=1}^{n}d_i\alpha_i\prod_{j=1}^{i-1}(1-\alpha_j),
\end{equation}
where $\mathbf{c_i}$ is the color computed from the SH coefficients of $i$-th Gaussian, and $\alpha_i$ is given by evaluating a 2D covariance matrix $\mathbf{\Sigma^{'}_i}$ multiplied by the opacity $o_i$.
The 2D covariance matrix is calculated by $\mathbf{\Sigma^{'}}=\mathbf{JW\Sigma W^{T}J^{T}}$, where $\mathbf{J}$ denotes the Jacobian of the affine approximation of the projective transformation, and $\mathbf{W}$ is the view transformation matrix.

\subsection{Holistic Gaussian Initialization}
\label{initialization}
The original 3DGS \cite{kerbl20233d} relies on SfM algorithms (mostly COLMAP \cite{schonberger2016structure}) to generate initialized points. However, we empirically find it is a time-consuming pipeline (minutes per scene, as shown in Sec. \ref{ablation}) for Gaussian initialization and tends to generate sparse points, leading to longer subsequent Gaussian optimization. Therefore, we delicately design a holistic Gaussian initialization strategy that can generate dense and accurate initialized points within seconds, and can also work for both binocular and monocular input image sequences. 

\noindent \textbf{Initialization with Binocular Input.} Given input binocular images $\{\mathbf{I_i^l, I_i^r}\}_{i=1}^{T}$, where $T$ refers to the time length, we first use the stereo depth estimation model \cite{Li_2021_ICCV} to predict the metric depth maps $\{\mathbf{D_i}\}_{i=1}^{T}$ of left views, following EndoNeRF \cite{wang2022neural}. Then, based on the predicted $\mathbf{D_i}$, we re-project pixels of each left image $\mathbf{I_i^l}$ into the world coordinates, obtaining the partial point cloud $\mathbf{P_i}$:
\begin{equation}
\label{Eq:reprojection}
    \mathbf{P_i}=\mathbf{K^{-1}}\mathbf{T_i}\mathbf{D_i}(\mathbf{I_i}\odot{\mathbf{M_i}}),
\end{equation}
where $\odot$ refers to the element-wise product, the binary mask $\mathbf{M_i}$ is used to filter out surgical tool pixels, and $\mathbf{K}$ and $\mathbf{T_i}$ refer to the known camera intrinsic and extrinsic parameters, respectively. Considering a single image contains a limited perspective and some tissue regions are occluded in the current view, we combine all the re-projected point clouds to achieve a holistic initialization:
\begin{equation}
\label{Eq:combinition}
    \mathbf{P}= \{\mathbf{P_1}, \mathbf{P_2}, ..., \mathbf{P_T}\}
\end{equation}

\noindent \textbf{Initialization with Monocular Input.} Given input monocular image sequence $\{\mathbf{I_i}\}_{i=1}^{T}$ we use the recent monocular depth estimation model \cite{depthanything} to predict the relative depth maps $\{\mathbf{D_i}\}_{i=1}^{T}$. Then, similar to the binocular input sequence, we also utilize the re-projection in Eq. \ref{Eq:reprojection} and combination in Eq. \ref{Eq:combinition} to obtain holistic initialized points. It is worth mentioning that though the relative depth maps lose scale information, we can still achieve accurate reconstruction as the Gaussian optimization process incorporates explicit geometric constraints from real poses and rendered images.

\subsection{Spatio-temporal Gaussian Tracking}
\label{tracking}
To model surface dynamics that are prevalent in the surgical procedure, we delicately design a deformation field $\mathbf{D}(\boldsymbol{\mu}, t)$ to track the attribute shift $\Delta \textbf{G}$ of each Gaussian at time $t$, based on which the deformed Gaussians $\mathbf{G_t}=\mathbf{G_0}+\Delta \mathbf{G}$ can be computed to render images. One feasible design is to use large neural networks to approximate $\mathbf{D}(\boldsymbol{\mu}, t)$, yet we empirically find this would incur slow inference speed and sub-optimal optimization (Sec. \ref{ablation}). Therefore, we instead split the deformation field into two lightweight modules $\mathbf{D}=\mathbf{F}\circ\mathbf{E}$, where $\mathbf{E}$ is a decomposed encoding voxel and $\mathbf{F}$ denotes Gaussian deformation decoder.

\noindent \textbf{Decomposed Encoding Voxels.} The encoding voxel $\mathbf{E}(\boldsymbol{\mu}, t)$ is used to encode the 4D inputs, i.e., the center of each Gaussian $\boldsymbol{\mu}$ and time $t$, into the time-aware latent feature $\mathbf{f}$. Inspired by \cite{wu20234d,yang2023neural}, we represent the 4D structural encoder as a multi-resolution HexPlane \cite{cao2023hexplane}, where the 4D encoding voxel $\mathbf{E}$ are decomposed as six planes with corresponding vectors:

\begin{equation}
\label{Eq:decomposition}
\mathbf{E}=\mathbf{E^{XY}}\otimes\mathbf{E^{ZT}}\otimes\mathbf{v^1}+\mathbf{E^{XY}}\otimes\mathbf{E^{ZT}}\otimes\mathbf{v^2}+\mathbf{E^{XY}}\otimes\mathbf{E^{ZT}}\otimes\mathbf{v^3},
\end{equation}

\noindent where $\otimes$ refers to the outer product, $\mathbf{E^{AB}}\in\mathbb{R}^{AB}$ is a learned plane of features, and $\mathbf{v^i}\in\mathbb{R^{D}}$ denotes the feature vector along $i$-th axis. To query a latent feature $\mathbf{f}$ given the continuous inputs $(x,y,z,t)$, we project the 4D coordinates onto the decomposed 2D planes and use the bilinear interpolation to compute features of each plane, finally obtaining the latent feature $\mathbf{f}$ through Eq. \ref{Eq:decomposition}. Through such decomposition, the computational cost is reduced from $\mathcal{O}(N^3)$ to $\mathcal{O}(N^2)$, which leads to a considerable acceleration of training and rendering speed. 

\noindent \textbf{Gaussian Deformation Decoder.} To decode the Gaussian deformation from latent $\mathbf{f}$, we design the decoder $\mathbf{F}$ as four tiny MLPs, $\mathbf{F}=\{\boldsymbol{F_\mu}, \mathbf{F_R}, \mathbf{F_S}\, \mathbf{F_o}\}$, to predict the deformation of position, rotation, scaling, and opacity of Gaussians, respectively. With the deformation of position $\Delta\boldsymbol{\mu}=\mathbf{F_\mu}(\mathbf{f})$, rotation $\Delta \mathbf{R}=\mathbf{F_R}(\mathbf{f})$, scaling $\Delta \mathbf{S}=\mathbf{F_S}(\mathbf{f})$, and opacity $\Delta o =\mathbf{F_o}(f)$, the deformed Gaussians $\mathbf{G_t}$ at time $t$ can be expressed as:
\begin{equation}
   \mathbf{G_t}=\mathbf{G_0}+\Delta \mathbf{G}=(\boldsymbol{\mu}+\Delta \boldsymbol{\mu}, \mathbf{R}+\Delta \mathbf{R}, \mathbf{S}+\Delta\mathbf{S}, o+\Delta o, SH),
\end{equation}
where the deformation of SH coefficients is not modeled, since modeling the position, rotation, scaling, and opacity are sufficient enough to capture the tissue movement and shape variations.

\subsection{Optimization}
\label{optimization}
Overall, the proposed framework is optimized by 1) rendering constraints to minimize the difference between the rendered and actual results and 2) spatio-temporal smoothness constraints on the rendering results.

\noindent \textbf{Rendering Constraints.} 
The rendering constraints consist of color rendering constraint $\mathcal{L}_{color}$ and depth rendering constraint $\mathcal{L}_{depth}^{B}/\mathcal{L}_{depth}^{M}$ for binocular/monocular input image sequence:
\begin{equation}
    \mathcal{L}_{color}=\sum_{\mathbf{x}\in\mathcal{I}}||\mathbf{M(x)}(\mathbf{\hat{C}(x)}-\mathbf{C(x))}||_1,\mathcal{L}_{depth}^{B}=\sum_{x\in\mathcal{I}}||\mathbf{M(x)}(\mathbf{\hat{D}^{-1}(x)}-\mathbf{D^{-1}(x))}||_1, 
\end{equation}
\begin{equation}
    \mathcal{L}_{depth}^{M}=1-Cov(\mathbf{\mathbf{M}\odot \hat{D}}, \mathbf{M}\odot \mathbf{D})/\sqrt{Var(\mathbf{M}\odot \mathbf{\hat{D}})Var(\mathbf{M}\odot \mathbf{D})}
\end{equation}
where $\mathbf{M}$, $\{\mathbf{\hat{C}}, \mathbf{\hat{D}}\}$, $\{\mathbf{C},\mathbf{D}\}$, and $\mathcal{I}$ are binary tool masks, predicted colors and depths using Eq. \ref{Eq:render}, real colors and depths, and 2D coordinate space, respectively. It is worth noting that for binocular depth rendering constraint $\mathcal{L}_{depth}^{B}$, we take the reciprocal of depth maps for loss computation to ensure optimization stability. While for monocular depth rendering constraint $\mathcal{L}_{depth}^{M}$, we use the soften constraint to allow for the alignment of depth structure without being hindered by the inconsistencies in
absolute depth values.

\noindent \textbf{Spatio-temporal Constraints.} We adopt total variation (TV) losses to regularize the rendering results. To avoid black/white holes in the regions that are occluded by surgical tools, we use a spatial TV loss to constrain the predicted colors and depths: $\mathcal{L}_{spatial}=TV(\mathbf{\hat{C}})+TV(\mathbf{\hat{D}^{-1}})$. Similar to \cite{wu20234d}, we also adopt a temporal TV term $\mathcal{L}_{temporal}$ to constrain the encoding voxels.

\noindent \textbf{Final Objectives.}
The overall objective is established by combining the above terms:
\begin{equation}
    \mathcal{L}=(\lambda_1 \mathcal{L}_{color} + \lambda_2 \mathcal{L}_{depth}^{B}/\mathcal{L}_{depth}^{M}) + (\lambda_3 \mathcal{L}_{spatial}+\lambda_4 \mathcal{L}_{temporal}),
\end{equation}
where $\lambda_{i=1,2,3,4}$ are balancing weights.

\section{Experiments}
\subsection{Experiment settings}
\noindent \textbf{Datasets and evaluation}
We conduct experiments on two publicly available datasets, ENDONERF \cite{wang2022neural} and SCARED \cite{allan2021stereo}. ENDONERF \cite{wang2022neural} contains two cases of in-vivo prostatectomy data captured from stereo cameras at a single viewpoint, encompassing challenging scenes with non-rigid deformation and tool occlusion. SCARED \cite{allan2021stereo} collects RGBD images of five porcine cadaver abdominal anatomies, using a DaVinci endoscope and a projector. Following previous work \cite{zha2023endosurf}, we split the frame data of each scene into 7:1 training and testing sets. We evaluate our method by comparing it with recent surgical scene reconstruction methods \cite{wang2022neural,zha2023endosurf,yang2023neural} using standard image quality metrics following \cite{wang2022neural}, including PSNR, SSIM, and LPIPS. Additionally, we record the training time, inference speed (FPS, frames-per-second), and GPU storage used for training.

\begin{figure}[t]
 \centering
  \includegraphics[width=1.0\linewidth]{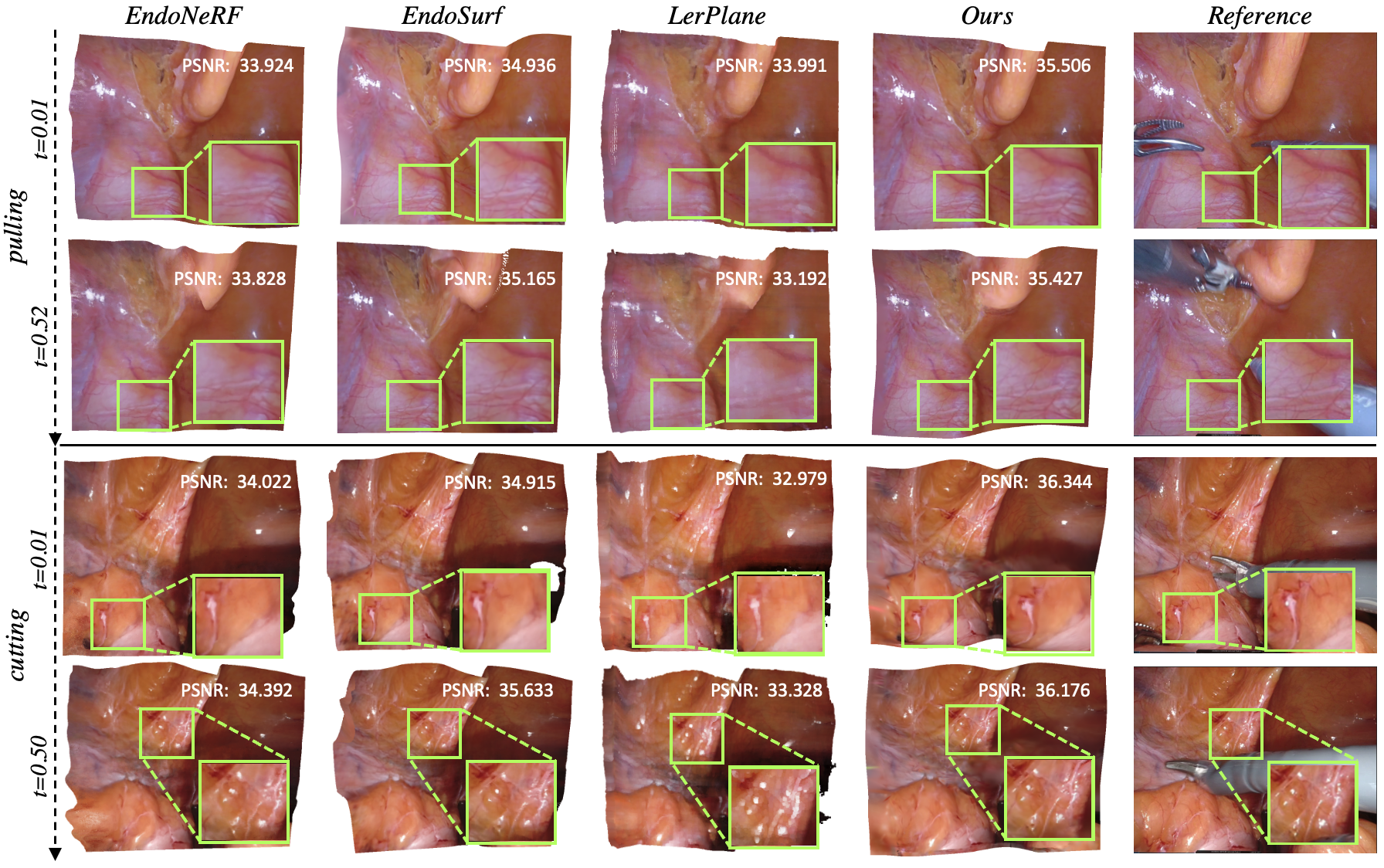}
 \caption{Illustration of the rendered images of previous works and ours.}
 \label{fig:comparison}
\end{figure}

\noindent\textbf{Implementation details}
In the initialization stage, we randomly sample $0.1\%$ points to reduce the redundancy. We use Adam \cite{kingma2014adam} as the optimizer with an initial learning rate $1.6\times 10^{-3}$.
A warmup strategy is used to first optimize Canonical Gaussians for 1k iterations, and then optimize the whole framework for 3k iterations. All experiments are conducted on a single RTX 4090 GPU and Intel(R) Xeon(R) Gold 5418Y CPU, using pure PyTorch framework \cite{paszke2017automatic}.

\subsection{Main results}
We compare our proposed method against existing SOTA reconstruction methods: EndoNeRF \cite{wang2022neural}, EndoSurf \cite{zha2023endosurf}, and LerPlane \cite{yang2023neural}. As shown in Tab. \ref{results}, we observe that EndoNeRF and EndoSurf achieve high-quality reconstruction of deformed tissues but require hours for optimization, which is quite computationally expensive. In contrast, LerPlane-9k greatly accelerates the training process to only around 3 minutes, yet compromises the reconstruction performance. More iterations of Lerplane-32k can promote the rendering quality of LerPlane, but it still suffers from slow inference speed. Our method EndoGaussian (binocular), on the other hand, achieves state-of-the-art reconstruction results of (37.849 PSNR) using only 2 minutes of training on the ENDONERF dataset, and most importantly, achieves a real-time rendering speed of around 195 FPS, providing more than $100\times$ acceleration over existing methods. Moreover, we observe our method only requires 2GB GPU memory for optimization, which is around 10$\times$ less than previous methods, releasing the hardware requirement when deploying in surgical practice. In addition, we observe using monocular input sequences, our method can also present promising rendering results with real-time rendering speed, revealing the generality of our method. To provide intuitive comparisons, we also illustrate several qualitative results in Fig. \ref{fig:comparison}. It can be observed that our method can preserve more details and provide better visualizations of the deformable tissues compared to other methods. These results demonstrate that EndoGaussian achieves real-time and high-quality surgical scene reconstructions, showing significant promise for future intraoperative applications.  

\subsection{Ablation Study}
\label{ablation}
\noindent \textbf{Gaussian Initialization} We experiment with `random' initialization that randomly generates initialized points and `COLMAP' initialization that produces sparse point clouds. From Tab. \ref{tab:ablation}, we observe that `random' greatly hinders model optimization and leads to poor reconstruction results, while initializing from `COLMAP' can lead to acceptable reconstruction results, it suffers from quite long initialization time and slow inference speed. In contrast, our initialization method introduces negligible time cost while maintaining better reconstruction quality and faster training and rendering speed.\\
\noindent{\textbf{Gaussian Tracking}}
\label{ablation-deform}
The encoding voxel proposed in Sec. \ref{tracking} can encode
Gaussians’ spatio-temporal information with minor optimization and rendering burden. As shown in Tab. \ref{tab:ablation}, replacing it with MLPs gives worse rendering quality, optimization time, and rendering speed, as MLPs have no spatio-temporal priors as HexPlane \cite{cao2023hexplane} and also introduce more optimization and rendering burden. 

\begin{table}[tbp]
\scriptsize
\centering
\setlength{\tabcolsep}{2pt}
\caption{Performance comparison on the ENDONERF \cite{wang2022neural} and SCARED \cite{allan2021stereo} dataset.
}
\label{results}
\begin{tabular}{c|c|ccc|ccc}
\toprule
Dataset & Method & PSNR$\uparrow$ & SSIM$\uparrow$ & LPIPS$\downarrow$ & TrTime$\downarrow$ & FPS$\uparrow$ & GPU$\downarrow$\\
\hline
\multirow{6}*{ENDONERF} & EndoNeRF \cite{wang2022neural}
& 36.062 & 0.933 & 0.089 & 5.0 hours & 0.04 & 19GB\\
& EndoSurf \cite{zha2023endosurf} & 36.529 & \underline{0.954} & 0.074 & 8.5 hours & 0.04 & \underline{17GB}\\
& LerPlane-9k \cite{yang2023neural} & 34.988 & 0.926 & 0.080 & \underline{3.5 min} & \underline{0.91} & 20GB\\
& LerPlane-32k \cite{yang2023neural} & \underline{37.384} & 0.950 & \textbf{0.047} & 8.5 min & 0.87 & 20GB\\
& Ours-monocular & 36.429 & 0.951 & 0.089 & \textbf{2.0 min} & \underline{180.06} & \textbf{2GB} \\
& Ours-binocular & \textbf{37.849} & \textbf{0.963} & \underline{0.054} & \textbf{2.0 min} & \textbf{195.09} & \textbf{2GB}\\
\midrule
\multirow{3}*{SCARED} & EndoNeRF \cite{wang2022neural} & 24.345 & 0.768 & \underline{0.313} & 3.5 hours & 0.02 & 22GB \\
& EndoSurf \cite{zha2023endosurf} 
& \underline{25.020} & \underline{0.802} & 0.356 & 5.8 hours & 0.01  & 22GB \\
& Ours-monocular & 23.477 & 0.744 & 0.489 &  \underline{5.01 min} & \underline{175.63} & \underline{3GB}\\
& Ours-binocular & \textbf{27.042} & \textbf{0.827} & \textbf{0.267} & \textbf{2.15 min} & \textbf{181.20} & \textbf{2GB}\\
\bottomrule
\end{tabular}
\end{table}

\begin{table}[tbp]
\scriptsize
\centering
\setlength{\tabcolsep}{2pt}
\caption{Ablation study of the designed components on ENDONERF \cite{wang2022neural}.}
\label{tab:ablation}
\begin{tabular}{c|c|ccc|ccc}
\toprule
Component & Method & PSNR$\uparrow$ & SSIM$\uparrow$ & LPIPS$\downarrow$ & InitTime$\downarrow$ & TrTime$\downarrow$ & FPS$\uparrow$ \\
\hline
\multirow{3}*{Gaussian initialization } & Random
& 6.023 & 0.282 & 0.604 & 0.1 sec& 2.0 min & 197.78\\
& COLMAP & 35.201 & 0.952 & 1.065 & 6.0 min & 4.0 min & 60.28 \\
& Ours & 37.849 & 0.963 & 0.054 & 2.0 sec& 2.0 min & 195.09 \\

\midrule
\multirow{2}*{Gaussian tracking} & MLP
& 34.834 & 0.936 & 0.095 & 2.0 sec & 9.0 min & 144.32 \\
& Ours & 37.849 & 0.963 & 0.054 & 2.0 sec& 2.0 min & 195.09 \\

\bottomrule
\end{tabular}
\end{table}

\section{Conclusion}
In this paper, we propose a real-time and high-quality framework for dynamic surgical scene reconstruction. By utilizing Holistic Gaussian Initialization and Spatio-temporal Gaussian Tracking, we can handle non-trivial Gaussian initialization and tissue deformation problems. Comprehensive experiments show that our EndoGaussian can achieve state-of-the-art reconstruction quality with real-time rendering speed, which is over $100\times$ faster than previous methods. We hope the emerging Gaussian Splatting-based reconstruction techniques could inspire new pathways for robotic surgery scene understanding, and empower various downstream clinical tasks, especially intraoperative applications.

\bibliographystyle{splncs04}
\bibliography{reference}

\end{document}